# Statistical Texture Features based Handwritten and Printed Text Classification in South Indian Documents

Mallikarjun Hangarge[a,*] K.C. Santosh[b,*] Srikanth Doddamani[a], Rajmohan Pardeshi[a]

[a]Department of Computer Science
Karnatak Arts, Science and Commerce College, Bidar, Karnataka, India
[b]INRIA Nancy-Grand Est, LORIA, BP - 239, 54506
Vandœuvre-lès-Nancy Cedex, France

**Abstract**

In this paper, we use statistical texture features for handwritten and printed text classification. We primarily aim for word level classification in south Indian scripts. Words are first extracted from the scanned document. For each extracted word, statistical texture features are computed such as mean, standard deviation, smoothness, moment, uniformity, entropy and local range including local entropy. These feature vectors are then used to classify words via k-NN classifier. We have validated the approach over several different datasets. Scripts like Kannada, Telugu, Malayalam and Hindi i.e., Devanagari are primarily employed where an average classification rate of 99.26% is achieved. In addition, to provide an extensibility of the approach, we address Roman script by using publicly available dataset and interesting results are reported.

## 1. Introduction

Advance development in digital technologies demands a robust Optical Character Recognition (OCR) system to reach the dream of paperless office. An overall performance of OCR system depends on the preprocessing technique such as segmentation, for instance. Among many, one of the prominent techniques is to separate handwritten and printed text, from a document image - a challenging problem in document image analysis domain, commercially speaking. In the context of Indian documents, without a surprise there exist more than two different scripts in a single document such as application forms, question papers, handwritten annotated documents and bank cheques (*cf.* Fig. 1).We repeat that handwritten and printed text separation is not a new domain of research. Besides Roman, Japanese, Chinese, Greek and Arabic scripts, a very few works provide attention to two major Indian scripts named Devanagari and Bangla, in particular. In India there are 12 scripts and 22 official languages are in use. In this paper, we attempt to deal with major South Indian scripts such as Kannada, Telugu and Malayalam, using statistical texture features and k-NN classifier. To realise the approach as a script independent handwritten and machine printed text separation, we report another experimental test over Roman script.

### 1.1. Related Work

In [1], a well-known technique is employed to compute histogram of width, height, gap and centre distance of extracted connected components. A polynomial classifier is used for classification of data fields, to decide that whether the data field is handwritten or printed. Another method is presented using a layered feed forward neural network [2]. It uses a set of features via histograms of gradient vector directions and luminance levels of gray scale image. The scheme classifies handwritten character, printed character, photograph and images in document image. However, it does not explicitly explain in those cases when handwritten and printed text zones or characters are intersected or overlapped. In [3], authors computed straightness of horizontal and vertical oriented lines and symmetric relative different points and these were used to discriminate between handwritten and printed character

---

using feed forward neural network. Violante et al., 1955 reported a scheme for separation of handwritten and printed address blocks of envelopes [4]. In particular, region, edge straightness, horizontal profile and dimensions of address box such as area, height, and width are computed. Classification task is carried out using multilayer preceptron neural network. Pal et al., 2001 [5] developed an algorithm for identification of machine printed and handwritten text line for two major Indian scripts: Devangari and Bangla using projection profile and statistical features with tree classifier. Hidden Markov Models (HMMs) with projection profiles are employed to separate handwritten text form machine printed text in document images on a word level basis [6]. In [7], authors are focused on structural features, a co-occurrence histogram, $2 \times 2$ grams, pseudo run lengths and Gabor filters. Fisher classifier is used for classification and Markov Random Fields (MRFs) are used for post processing to improve results. E. Kavallieratou et al., 2004 [8] developed a scheme for classification of handwritten and printed Greek text using structural features and discriminant analysis. In some cases, Gabor filters are used for feature extraction and probabilistic neural networks for classification of handwritten and printed text in Arabic document images [9]. Features such as shape characteristics are dominantly used for handwritten and printed text separation via k-NN classifier [10, 11]. In [12], a system using chain code features and support vector machine (SVM) classifier is reported where texts are presented as sparse content and arbitrary oriented document fragments. As a reminder, computing chain code (for 8 different directions) is very expensive in case the boundary of the characters is not smooth enough. The Radon transform has also been used to represent text that goes word-wise together with SVM for classification [13]. As reported in [14], Eigen-faces features have been popularly used for text separation while defining the local threshold.

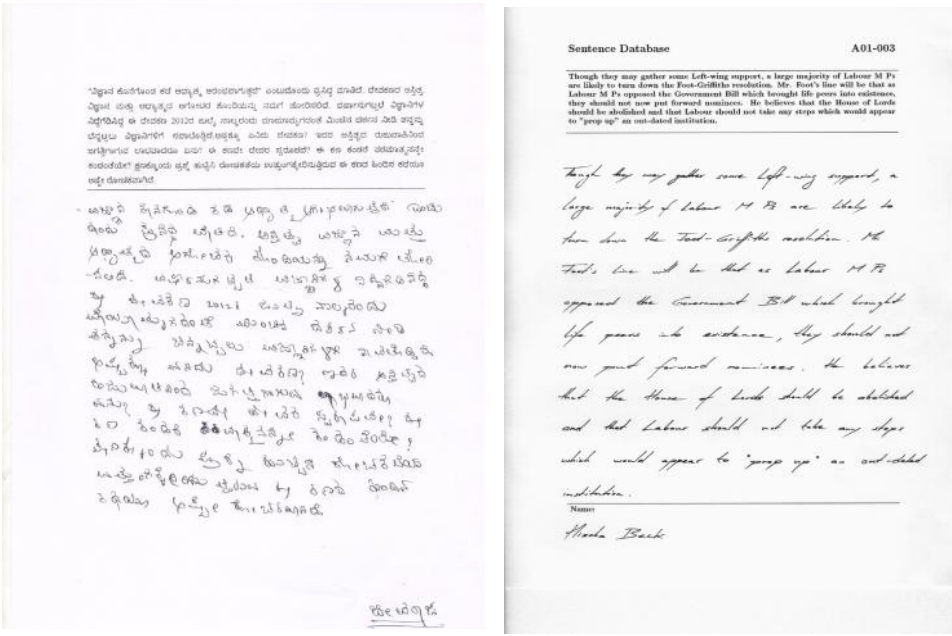

Fig. 1. A couple of samples showing each document containing handwritten and printed texts.

Based on our quick review made in the aforementioned paragraph, without a surprise, one can notice that several different works use script dependent features and these may not accommodate in case when new scripts are included. This in primarily due to the fact that shape of the characters of one script wills differ from another. Besides, the reported woks are addressed the problem in different ways like character level [1], word level [10, 11, 12] and bock/line level separation. However, one cannot not cleanly judge which one is really convincing. Furthermore, it is important to repeat that the approaches that deal with the text separation problem are application dependent. Considering our context i.e., Indian scripts, word level handwritten and printed text separation would be interesting choice since there exists texts are interlaced in word level.

*1.2. Structure of the paper*

The paper is organised as follows. We start with detailing the proposed approach in Section 2. It mainly includes pre-processing i.e., word level segmentation in Section 2.1, feature computation in Section 2.2 and classification in Section 2.3. Full experiments are reported in Section 3 and analyse the approach in accordance with the scripts employed. We mainly include a comprehensive experimental result analysis. We conclude our paper in Section 4.

**2. Proposed Approach**

We primarily aim for word level classification in south Indian scripts. To handle it, words are first extracted from the scanned document. For each extracted word, statistical texture features are computed. These features are integrated to form a single feature vectors which are then used for handwritten and printed text separation via k-NN classifier. For better understanding, we refer readers to Fig. 2 where it provides a schematic work-flow of the system.

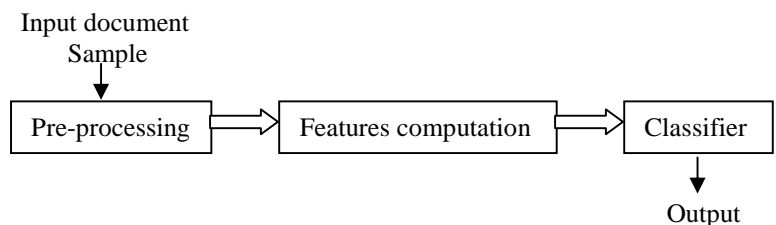

Fig. 2. A screen-shot of an overall work-flow of the system.

*2.1. Preprocessing*

The preliminary task is to do pre-processing. Pre-processing techniques are application dependent. In our case, we aim for segmenting words from the scanned documents. As an example, Fig. 1 shows a couple of sample images where handwritten and printed texts are shown in a single document. To handle it, it consists of a three-step process. We first binarised the document using Otsu's threshold selection method [15] and employed basic morphological operators to remove outliers such as commas, semicolons, single and double quotations. Similarly, long lines are removed based on connected component analysis. Word segmentation is then applied by using dilation so that words separation is possible where connected component rule can basically be applied. In Fig. 3, for visual understanding, we provide bounding boxes to all words taken from a sample document.

*2.2. Features Computation*

The problem of handwritten and printed text separation is considered as a problem of texture analysis. Human perception on discriminating between handwritten and printed text is exploiting the texture properties of the text. Thus, we take advantage of texture features for feature extraction. Texture features are first reported in [16] for image classification. For better understanding, texture can also be defined as: it is property which contains important information about structural arrangement of surfaces and their relationship with surrounding environment. In this paper, we use simple statistical measure of texture from image, which observe texture as quantitative measure of arrangement of intensity level in image region. Statistical properties of the intensity histogram such as (F1) mean, (F2) standard deviation, (F3) smoothness, (F4) third moment, (F5) uniformity, (F6) entropy, (F7) neighborhood standard deviation, (F8) local range and (F9) local entropy. These set of nine statistical texture features collectively used to generate a feature vector. We have calculated these statistical properties using statistical moments.

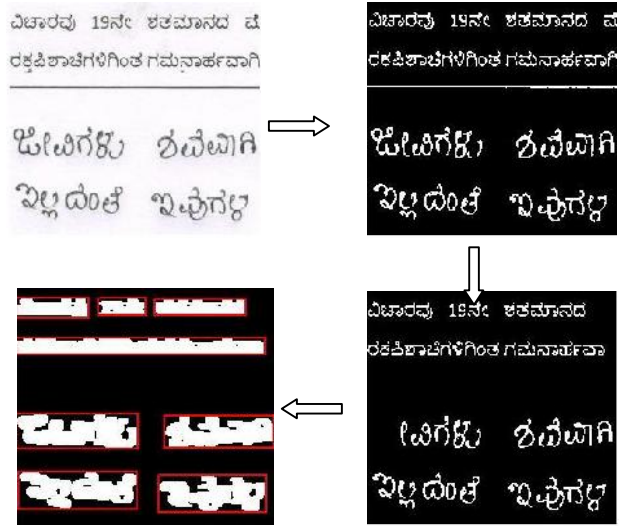
Fig.3. An example showing word segmentation using Kannada script.

Let $r_i$ be the discrete random variable which denotes the intensity levels in image and $h(r_i)$; $i = [0; 1; 2; \ldots ; k-1]$, be the corresponding normalised histogram, where k is the possible intensity value. To describe the shape of histogram with principal approach is via its central moment and the expression of $n^{th}$ moment for mean can be expressed as follows,

$$\mu_i = \sum_{i=0}^{K-1} (r_i - m)^n \tag{1}$$

In the following, the above stated nine features are defined and calculated for a given n.

1) F1 is a measure of average intensity, used for calculation of average intensity of word image,
$$m = \sum_{i=0}^{k-1} r_i\, p(r_i) \tag{2}$$

2) F2 is taken as measure of average contrast and is expressed as,
$$\sigma = \sqrt{\sigma^2} \tag{3}$$

3) F3 measures the relative smoothness of the intensity in a region which is defined as,
$$R = 1 - \frac{1}{1+\sigma^2}. \tag{4}$$

4) F4 measures the skewness of an intensity histogram which can be expressed as,
$$\mu_3 = \sum_{i=0}^{K-1} (r_i - m)^3 h(r) \tag{5}$$

5) F5 measures the uniformity of pixels which is defined as,
$$U = \sum_{i=0}^{K-1} (p)^3(r_i) \tag{6}$$

6) F6 computes the entropy i.e., randomness of pixels in image and is expressed as,
$$e = \sum_{i=0}^{K-1} p(r_i) \log_2 p(r_i) \tag{7}$$

7) F7 computes local standard deviation of an image.
8) F8 determines local range of an image.
9) F9 determines local entropy of an image.

## 2.3. Classification

The traditional and simplest classification algorithm is *k*-nearest neighbour algorithm (*k*-NN). It is a method of classifying the instances based on the nearest training examples in the feature space. It classifies an object based on a majority vote of its neighbours, with the object being assigned to the class most common amongst its *k* nearest neighbours. The best selection of *k* value depends on the nature of the data under classification. However, larger value of *k* reduces the effect of noise on the classification, but increase the time complexity. It eventually gives weak boundaries between the classes. The smaller value of *k* has high influence of noise.

## 3. Experiments

*3.1. Datasets*

Due to the unavailability of data set, we have created a dataset of 5000 printed and 5000 handwritten word images. A sheet of A4 size, which contains few lines of machine printed text at the top and the same, is provided to 100 writers of different professions. Then, writers are asked to write the same text in the blank space provided below the printed text region. These 100 A4 size pages were digitised by scanning at 300 dpi. By applying segmentation algorithm as discussed in Section 2.1, obtained a dataset of 1000 printed and 1000 handwritten text word images of each script. For English, IAM DB 3.0 dataset is used and extracted 1000 handwritten and 1000 printed text words. In this way, a dataset of size 2000 word images of each script namely Kannada, Telugu, Tamil, Malayalam and Roman were created. Further, 5000 printed and 5000 handwritten text words of five scripts are mixed to generate a multilingual dataset. A few sample of multilingual dataset is shown in Fig. 4.

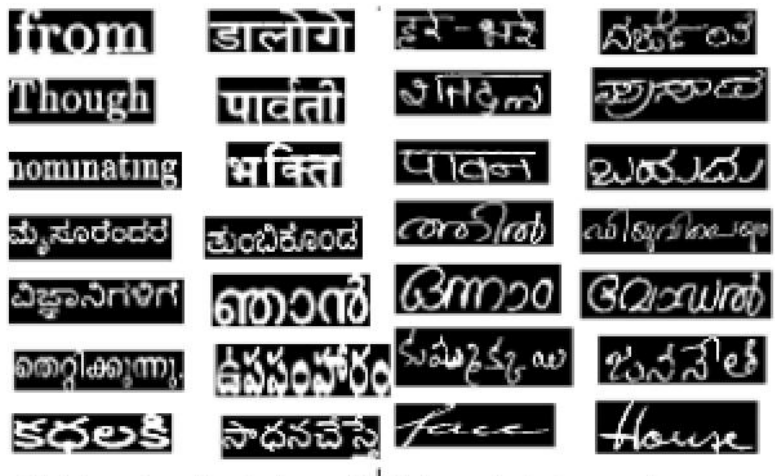

Fig.4. A sample multi-script dataset of handwritten and printed text words.

*3.2. Evaluation Protocol*

To evaluate the performance of the method, K-fold cross validation (CV) has been implemented unlike traditional dichotomous classification. In K-fold CV, the original sample for every dataset is randomly partitioned into K subsamples. Of the K sub-samples, a single sub-sample is used for validation and the remaining K = 1 sub-sample is used for training. This process is then repeated for K folds, with each of the K sub-samples used exactly once. Eventually, a single value results from averaging all. In our experimental tests, the value of K = 10.

Table 1. Recognition accuracy of handwritten and printed word separation using multilingual dataset for k-NN with k = 1,3,5,7.

|  | k = 1 | k = 3 | k = 5 | k = 7 |
|---|---|---|---|---|
| Handwritten | 98.92 | 98.90 | 99.00 | 98.92 |
| Printed | 99.54 | 99.54 | 99.52 | 99.50 |
| Average | 99.23 | 99.22 | 99.26 | 99.21 |

Table 2. Confusion matrix of handwritten and printed word separation using multilingual dataset for k = 5.

|  | Handwritten | Printed | Total |
|---|---|---|---|
| Handwritten | 4950 | 50 | 5000 |
| Printed | 24 | 4976 | 5000 |

Table 3. Script-wise recognition accuracy of handwritten and printed word separation for k = 5.

|  | Roman | Hindi | Kannada | Telugu | Malayalam |
|---|---|---|---|---|---|
| Handwritten | 97.40 | 100 | 98.20 | 100 | 100 |
| Printed | 98.60 | 100 | 98.90 | 100 | 100 |
| Average | 98.00 | 100 | 98.55 | 100 | 100 |

*3.3. Results and Analysis*

To observe the script independent behavior of our algorithm, a comprehensive study has been made from the thorough experimental tests that are conducted on multilingual word dataset. Having the dataset of 5000 handwritten and 5000 printed words from five scripts, our experimental tests will go like this.
1) Use multilingual dataset for different values of k and select an optimal value of it.
2) The selected value is then used for script-wise test. The second test attests the fact that which script does not show discriminant behavior between handwritten and printed text, considering the similar statistical texture features.

In Table 1, multilingual dataset is used for different values of k = 1, 3, 5and 7 in k-NN classifier. In this test, we have found an optimal performance of the k-NN classifier happens for k = 5. For deeper analysis, we have provided its confusion matrix in Table 2. As soon as we have received an optimal selection of the value of k, another experiment i.e., script-wise test has been made with it. Such a script-wise experimental results are provided in Table 3. Based on results reported in Table 3, tests over scripts like Hindi, Telugu and Malayalam showed promising performance, compared to Roman and Kannada scripts. We have found that false recognition of handwritten exist confusion of clean handwritings of any individuals with the printed ones and vice-versa. This type of confusion is occurred between Kannada handwritten and printed text.

## 4. Conclusion and Further work

In this paper, our study emphasises the simplicity and efficiency of the statistical texture features in separating south Indian scripts such as Kannada, Telugu, Malayalam and Hindi. It has been tested in terms of word level handwritten and printed text separation. The classification of scripts is carried on a large dataset of 10000 word images using k-NN algorithm. Basing on our reported results, the approach can be taken as script, font and writing style independent. We have also noted that the approach provides interesting results while using Roman script.

Further work will be devoted to state-of-the-art comparison using publicly available dataset. In addition, we keep continuing to make our datasets publicly available.

**Acknowledgements**

This work is carried out under the UGC sponsored minor research project (ref: MRP(S):661/09-10/KAGU013/UGCSWRO).

*Statistical Texture Features based Handwritten and Printed Text Classification in South Indian Documents*